\documentclass[conference]{IEEETran}
\IEEEoverridecommandlockouts
\usepackage{cite}
\usepackage{amsmath,amssymb,amsfonts}
\usepackage{algorithmic}
\usepackage[linesnumbered,ruled,vlined]{algorithm2e}
\usepackage{adjustbox}
\usepackage{optidef}
\usepackage{graphicx}
\usepackage{subfigure}
\usepackage{mathtools}
\usepackage{circledsteps}
\usepackage{makecell}
\usepackage{wrapfig,booktabs}
\usepackage{array}

\usepackage{graphicx}
\usepackage{textcomp}
\usepackage{xcolor}
\usepackage{multirow}
\usepackage{fbox}
\usepackage{url}
\SetKwInput{KwInput}{Input}                
\SetKwInput{KwOutput}{Output}              

\def\BibTeX{{\rm B\kern-.05em{\sc i\kern-.025em b}\kern-.08em
    T\kern-.1667em\lower.7ex\hbox{E}\kern-.125emX}}
    
\makeatletter

\def\ps@IEEEtitlepagestyle{%
    \def\@oddfoot{\mycopyrightnotice}%
    \def\@evenfoot{}%
}
\def\mycopyrightnotice{%
    {\footnotesize  Accepted in DCOSS 2022. Not a Camera Ready manuscript. DOI will be added soon.\hfill}
    \gdef\mycopyrightnotice{}
}



\makeatletter
\newcommand*\titleheader[1]{\gdef\@titleheader{#1}}
\AtBeginDocument{%
  \let\st@red@title\@title%
  \def\@title{%
    \bgroup\normalfont\large\raggedright\@titleheader\par\egroup
    \vskip1.5em\st@red@title}
}
\makeatother
\title{FrameHopper: Selective Processing of Video Frames in Detection-driven Real-Time Video Analytics
}

\titleheader{Accepted in DCOSS 2022. © 2022 IEEE.  Personal use of this material is permitted.  Permission from IEEE must be obtained for all other uses, in any current or future media, including reprinting/republishing this material for advertising or promotional purposes, creating new collective works, for resale or redistribution to servers or lists, or reuse of any copyrighted component of this work in other works.}

\begin{document}

\author{\IEEEauthorblockN{Md Adnan Arefeen\IEEEauthorrefmark{1},
Sumaiya Tabassum Nimi\IEEEauthorrefmark{2}, and Md Yusuf Sarwar Uddin\IEEEauthorrefmark{3}}
\IEEEauthorblockA{School of Computing and Engineering\\
University of Missouri-Kansas City, MO, USA\\
\IEEEauthorrefmark{1}aa4cy@mail.umkc.edu,
\IEEEauthorrefmark{2}snvb8@mail.umkc.edu,
\IEEEauthorrefmark{3}muddin@umkc.edu}}

\maketitle

\begin{abstract}
Detection-driven real-time video analytics require continuous detection of objects contained in the video frames using deep learning models like YOLOV3, EfficientDet, etc. However, running these detectors on each and every frame in resource-constrained edge devices is computationally intensive. By taking the temporal correlation between consecutive video frames into account, we note that detection outputs tend to be overlapping in successive frames. Elimination of ``similar'' consecutive frames (the same set of objects with slightly offset bounding boxes) will lead to a negligible drop in performance while offering significant performance benefits by reducing overall computation and communication costs. The key technical questions are, therefore, (a) how to identify which frames to be processed by the object detector, and (b) how many successive frames can be skipped (called \emph{skip-length}) once a frame is selected to be processed. The overall goal of the process is to keep the error due to skipping frames as small as possible. We introduce a novel error vs processing rate optimization problem with respect to the object detection task that balances between the error rate and the fraction of frames actually passed and processed. Subsequently, we propose an off-line Reinforcement Learning (RL)-based algorithm to determine these skip-lengths as a state-action policy of the RL agent from a recorded video and then deploy the agent online for live video streams. To this end, we develop FrameHopper, an edge-cloud collaborative video analytics framework, that runs a lightweight trained RL agent on the camera and passes filtered frames to the cloud/edge server where the object detection model runs for a set of applications. We have tested our approach on a number of live videos captured from real-life scenarios and show that FrameHopper processes only a handful of frames but produces detection results closer to the ``\emph{oracle}'' solution and outperforms recent state-of-the-art solutions in most cases.
\end{abstract}


\section{Introduction}
The wide-ranging deployment of video cameras in the current society makes it pervasive in nature. Video cameras are mainly used for surveillance systems and are typically found in areas such as playgrounds, shopping malls, streets, and offices. In most cases, the live streaming of videos is required for scene understanding that leads to automated analyzing of video frames for detecting interesting phenomena such as such as object tracking~\cite{bastani2020miris}, person identification~\cite{gu2020appearance,hou2020temporal,yang2020spatial}, face detection~\cite{zhang2020refineface}, object detection~\cite{tan2020efficientdet} and classification~\cite{liu2021ntire,kortylewski2020combining}. Recent advancement in  deep learning models has enabled automated processing of images and video frames in a high degree of accuracy. However, these deep learning models are usually too demanding in terms of their computation, storage, and memory requirements to be deployed in the constrained settings of the edge devices. Thus accurate and real-time analytics on video and image data using deep learning models has emerged as a challenging research problem. 

Arguably sending all these live video frames to a cloud server for processing is not cost-effective. To overcome this problem, lightweight frame filtering techniques are applied at the edge even to the camera itself to filter out frames~\cite{reducto,pakha2018reinventing,kang2017optimizing,wang2018bandwidth}. Several pieces of research highlight the frame filtering technique to propose a comparatively low-cost frame difference detector to filter the redundant frames based on a certain application~\cite{chen2015glimpse}. But the frame filtering techniques highly depend on the type of applications that is running at a user side for a specific purpose. Obviously, detecting pedestrian on videos~\cite{cai2015learning} is different than counting vehicles on a road~\cite{hsu2020traffic,Peng_2021_WACV,Bhardwaj_2019_CVPR,fan2018watching,trafficgorkem} in terms of filtering frames. While the former one is an example of rare-event detection~\cite{hamaguchi2019rare,pang2020self} where the rare events may occur after a long interval, the latter one necessitates to determine after how many frames one vehicle can be gone out of the camera scene assuming the camera can be a static one.

\begin{figure}
    \centering
    \includegraphics[width=0.8\linewidth]{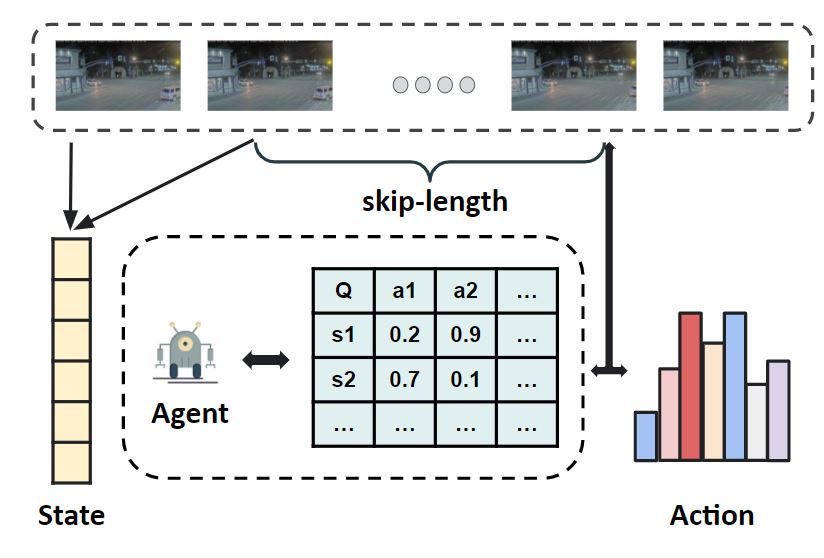}
    \caption{FrameHopper Overview}
    \label{fig:initial-model}
\end{figure}

Among different configurations~\cite{jiang2018chameleon} of video analytics pipeline, running object detection (labeling multiple objects in a video frame with their bounding boxes) is becoming a primitive operation for many video analytic applications~\cite{reducto, jiang2018chameleon, bastani2020miris}. But popular detection engines are very expensive to run. As consecutive video sequences for an interval contain temporal and spatial correlation, processing every frame results in a waste of resources associated with the video analytics pipeline. Central to any filtering or elimination of redundant frames is the trade-off between the number of frames being filtered and the amount of error or accuracy loss is introduced due to those filtering. These two are odd: we can filter more but that would result in higher error in upstream video processing pipelines, whereas less filtering may give less error but at the cost of higher processing rate and more computation down the line. A good filtering strategy needs to draw a rightful balance between these two.

In this paper, we propose a reinforcement learning-based lightweight real-time video frame filtering technique named \emph{FrameHopper}. Unlike earlier techniques of frame filtering such as Reducto~\cite{reducto} or FilterForward~\cite{filterforward}, FrameHopper neither buffer frames for a certain duration to apply the filtering nor process every incoming frame to detect whether it needs to be filtered or not; instead, it decides how many \emph{future} frames can be skipped based on the current frame that has just been processed and a small number of earlier processed frames. FrameHopper acquires this capability by analyzing an earlier recorded video (preferably from the same scene) and by observing the skipping patterns among those frames incommensurate with a certain degree of accuracy loss or performance degradation. To this effect, FrameHopper trains (in offline) a reinforcement learning (RL) agent that ``learns'' those skipping actions into a ``policy'' that effectively dictates on which situations (represented as ``states'') how many frames can be skipped (denoted as ``actions''). Once trained, the agent is applied in real-time to make skipping decisions on incoming video frames. The overview of FrameHopper is shown in Figure~\ref{fig:initial-model}. The overall contribution of the paper is as follows:
\begin{itemize}
    \item We propose a novel reinforcement learning-based lightweight frame filtering technique \emph{FrameHopper} to filter frames at the edge for accelerating the efficiency of detection-driven applications.
    \item Unlike recent frame filtering-based methods, the \emph{FrameHopper} skips the similar consecutive frames entirely at the edge reduces the computation cost of a frame queue for saving frames to filter out before sending them to the cloud.
    \item Furthermore, we also propose an automated threshold selection-based frame filtering method that will select a threshold that results in a minimal error due to skipping with a minimal fraction of processing frames.
    \item Evaluation on a number of real-time videos with respect to qualitative or quantitative comparison, FrameHopper acts closer to oracle solution and outperforms other state-of-the-art methods in most cases.
\end{itemize}


\section{Background and Challenges}
Deploying deep learning models in constrained settings of the edge devices due to the demand for high computation, storage, and memory requirements. Hence it is often the case that processing tasks like object detection are done in the remote servers (most often in the cloud) and the corresponding results are sent to the edge to reach the users. The main drawback of this approach is the transfer of a large amount of data between the cloud and the end device. This issue becomes more conspicuous for real-time processing jobs like traffic video surveillance, abnormal event detection, object tracking, etc. For this reason, edge computing has become the focus of research nowadays. Towards this end, a framework NestDNN~\cite{fang2018nestdnn} was proposed to make the resource-accuracy trade-offs dynamic where the multi-capacity model ensures the dynamic resource-accuracy trade-offs. In~\cite{jiang2018chameleon}, a configuration controller was designed that dynamically picks the best configuration for existing deep learning models for video analytics so that with the same amount of resources, the chosen configuration led to higher performance. Meanwhile, in Reducto~\cite{reducto}, frames were selected for processing based on temporal correlation between the features and by considering thresholds on different detection parameters to reduce the overhead associated with processing each frame at edge nodes, so that it was possible to obtain high accuracy with low latency. For live video analytics, a bandwidth-efficient architecture was proposed by~\cite{wang2018bandwidth}. To eradicate the total transmission, four different strategies were proposed. For real-time object recognition tasks at mobile devices, Glimpse~\cite{chen2015glimpse} offered an active cache that serves as an efficient frame filtering system for the reduction of network bandwidth. On the other hand, query-based video analysis is proposed in NoScope~\cite{kang2017noscope} where a binary classifier labeled a video frame true or false for a specified query. To determine the frame difference, a difference detector was implemented which measured the mean-squared distance of a given frame to a reference frame.

Offloading every video frame to the cloud for the queried applications that are running in the data center hampers performance in terms of latency in real-time streaming.
Hence a number of research works have been conducted towards this end. Filterforward~\cite{filterforward} is one such work that addressed this problem. Rather than offloading each of the frames, FilterForward only sent the relevant frames (filtering) to the cloud (forwarding) based on the applications running in the data-center and thus a moderate level of bandwidth was saved. Trained \emph{microclassifiers} (MC) were deployed at edge nodes for finding out relevant frames. Clowfish~\cite{nigade2020clownfish} proposes to deploy a comparatively lightweight but not so accurate detection model in the resource-constrained edge devices and deploy a the full model in the server. Through exploiting the spatio-temporal correlations between consecutive video frames, only a subset of the video frames were sent to the cloud for full processing. 
In the papers~\cite{zeng2020distream} algorithms were proposed for cross-camera video analytics.

Among lightweight frame filtering methods, either a lightweight neural network is considered or even less expensive frame features such as pixels are used for filtering frames. In spite of lightweight filtering, the earlier methods need to process each frame on edge to decide on filtering frames. In the case of FrameHopper, it hops from one frame to another without even computes on the intermediate frames between them based on the previous history which is learned through a reinforcement learning agent.

\section{Framehopper}\label{sec:framehopper}
In this section, we present \emph{FrameHopper}, a frame selection engine to achieve real-time object detection. We note that processing each and every video frame for detection in real-time streaming is redundant since consecutive frames mostly contain overlapping objects with a slight change in their locations. Hence in most cases, skipping a few consecutive frames still results in a very good detection performance with negligible error. But the extent to which we can skip frames cannot be a pre-determined choice; instead, it depends on the nature of the environment where the camera is deployed and the scene being observed. Skipping a \emph{constant} number of frames in all cases will lead to either a high degree of error or a high degree of redundancy due to uncertainty in the environment. This error vs redundancy is a fundamental trade-off, which resurface any real-time video processing pipelines. The goal of our developed FrameHopper is to introduce a dynamic frame sampling rate so that the frame processing time in a detection-driven environment is minimized by skipping the redundant frames entirely from processing.

As FrameHopper incorporates the notion of frame skipping, its first task is to identify a suitable \emph{skip-length}, the number of consecutive frames that can skipped after a certain frame is processed. 
Using this \emph{skip-length}, FrameHopper skips a number of consecutive frames and proceed. We, therefore, ask \emph{what is a good skip-length for any video given a certain environment in query-specific or query-agnostic scenario?} The FrameHopper algorithm answers this question by introducing a dynamic \emph{skip-length} based frame filtering approach. Precisely, FrameHopper answers the following questions.\\
\textbf{Q1} \emph{How to characterize the relationship between the fraction of processing frames and the error due to frame skipping?}\\
\textbf{Q2} \emph{How to define threshold-driven filtering without having the necessity of a threshold and how to identify a good skip-length with minimal training effort?}\\
\textbf{Q3} \emph{How to construct a lightweight solution that can map the simple frame difference algorithm to the object location information of a frame?}

\subsection{Objective}
The main goal of FrameHopper is to maximize the frame dropping rate and to minimize the error due to skipping. Let $N = \{f_1, f_2, \cdots, \}$ be the sequence of all frames in the video stream and we are to find a suitable subset $P \subseteq N$, by skipping some frames in $N$. Let $\kappa(i) = k$ (the skip-length) be the number of frames skipped when frame $f_i$ is selected to be processed ($f_i \in P$). The overall goal of the FrameHopper is to find: 
\begin{equation}
\min_{P\subseteq N} \sum_{f_i\in P} E(f_i,\kappa(i)) + \lambda |P| 
    \label{eqn:1}
\end{equation}
\noindent
where $E(f_i,\kappa(i))$ is the amount of \emph{error} when $k$ successive frames are skipped right after $f_i$. Conceptually, the amount of error due to skipping frames can be measured as the degree of ``difference'' between the frame that was last processed (the reference frame) and the successive frames that were skipped from the current reference frame up until the next reference frame (this is as if the same reference frame is replaced in the sequence as surrogate frames in place of the frames that were skipped). 
Ideally, when no frames are skipped (i.e., $\kappa(i)=0$), the error is zero, and the error grows monotonically with $k$ indicating that as more frames are skipped, more error is accumulated. With this, the error term can be defined as a cumulative sum of distances between frames:
\begin{equation}
 E(f_i,k) = \sum_{j=1}^{k} \mathcal{D}(f_i,f_{i+j})
    \label{eqn:2}
\end{equation}
Note that $E$ and $P$ are at odd: we can reduce the size of $P$ by skipping more frames (equivalently processing fewer frames) but at the expense of introducing higher error (E) due to skipping frames, and vice versa. The parameter $\lambda$ in Eqn~\ref{eqn:1} acts as a balancing factor between the two.
 
\subsection{Determining the Frame Difference}\label{sec:FrameDiff}
To consider the two frames to be similar from a detection result perspective, 
usually, the F1 score~\cite{chen2015glimpse,reducto} is the most popular metric to compute similarity between two frames based on their detection outputs. To compute the F1 score, the Intersect over Union (IoU) is calculated to measure the degree of overlap between bounding boxes between the two frames. If a pair of boxes (one from each frame) with the same class label has an IoU value exceeding a certain threshold, it is considered a ``hit'' (the threshold is usually set to 0.5). Based on the hits, precision and recall values are calculated and their harmonic mean gives the F1 score between the two frames in comparison. The higher the F1 score, the more similar the two frames are. The formal equations are given below:
\begin{equation}
    \begin{split}
    precision &=  \frac{\# \;of\;hits} {\#\; of\; detected\; objects} \\
    recall &= \frac{\# \;of\;hits} {actual \#\; of\; detected\; objects} \\
    F1 &= 2 \times \frac{precision\times recall}{precision + recall}\\
    \end{split}
\end{equation}
Consequently, we define the difference/distance function between two frames as $\mathcal{D}(f_i, f_j) = 1 -  F1 (f_i,f_j)$. The higher the F1 score, the lower the $\mathcal{D}$ between two frames.

Given a distance function between frames, our task is to find the best subset of frames as $P$ (out of $N$) that would minimize both error and the fraction of frames to be processed (as per Eqn~\ref{eqn:1}). 
But, Eqn~\ref{eqn:1} is intractable and NP-Hard for a given $N$ (the problem can be reduced to a partition problem) and not feasible to run it in online fashion. 
We consequently recognize that finding the skip-lengths on an incoming stream of frames can be posed as a \emph{sequential decision} problem that can be represented as a Markov Decision Process (MDP) and, therefore, can be solved by a reinforcement learning(RL) algorithm.
In the next section, we describe an RL-based FrameHopper algorithm and discuss how we incorporate the distance function $\mathcal{D}$ in building up the RL-Agent.

\subsection{Algorithm Design}
\begin{figure}
    \centering
    \includegraphics[width=\linewidth]{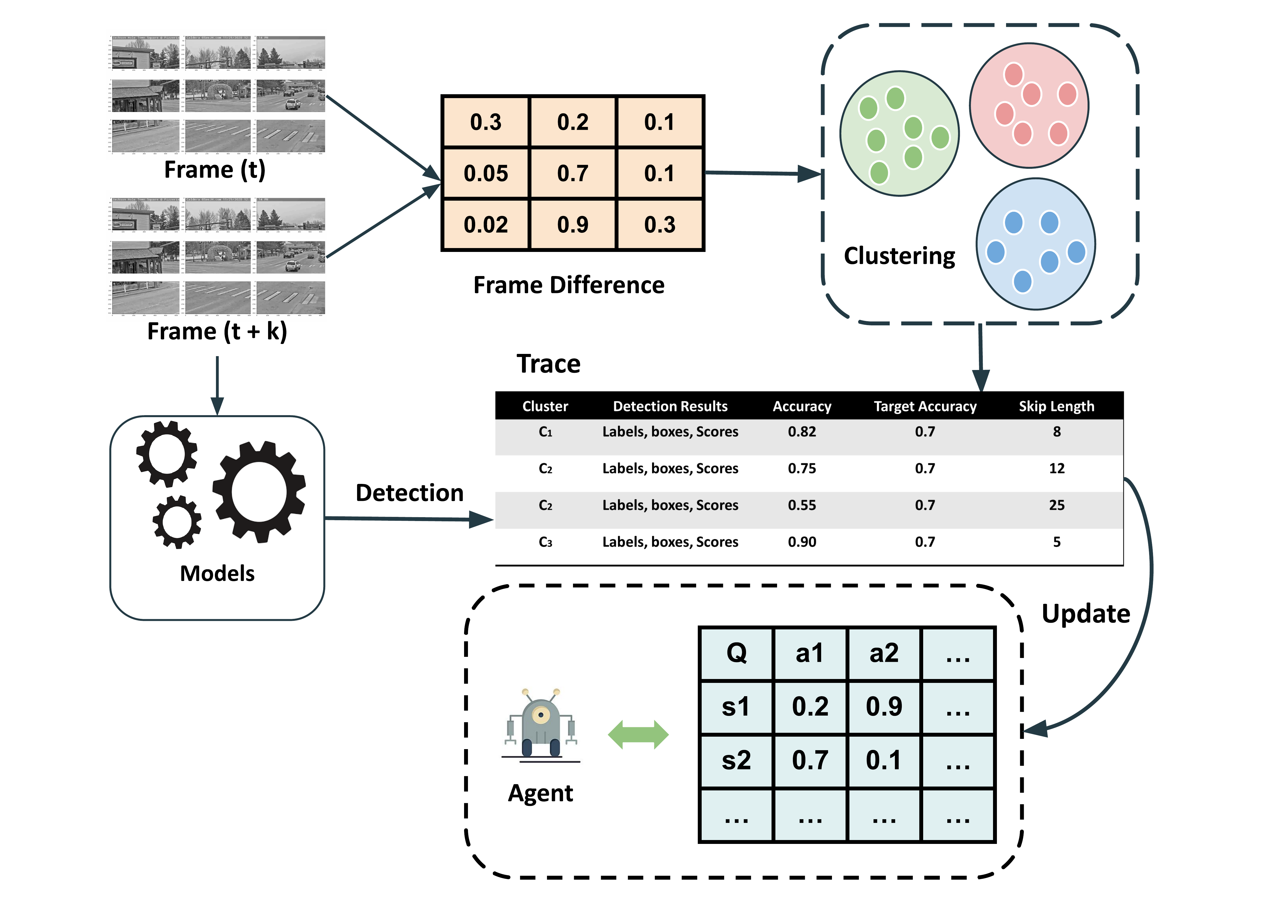}
    \caption{FrameHopper Agent Training Scheme.}
    \label{fig:agent-training}
\end{figure}


FrameHopper trains, in \emph{offline}, a reinforcement learning (RL) agent that replays an earlier recorded video and observes which frames can be suitably skipped in certain situations (encoded as ``states'') that results in a certain amount of bounded error due to the skipping of those frames (the error bound is a user defined parameter, described later on). The suitability of an action (that is, skipping a certain number of consecutive frames) on a given state depends the degree of error that the chosen action introduces, which is reflected in the associated ``reward'' value for that action. With this, the agent effectively learns those skipping actions into a ``policy'' that indicates the suitable skip-length at a certain state. This policy is essentially populated and stored as a state-action table (known as Q-table), which maps a state to a set of possible actions and their associated reward values. The objective of the RL training is to find a policy $\pi$ that tries to maximize the rewards in the long run. More formally, under a policy $\pi$, the Q-value for a given state-action pair $(s, a)$ will eventually converge to the following (with usual RL notations) :
\begin{equation}
    \mathbf{Q}_\pi(s,a) = \mathbb{E}_\pi\left[\sum_{k=0}^\infty \gamma^k\mathbf{R}_{t+k+1} | S_t=s,A_t=a\right]
\end{equation}

Once trained, the agent is deployed in real-time to make skipping decisions on incoming video frames. More formally, the agent determines its current state $s$ and takes action $a^* = \arg \max_a Q(s, a)$ (that is, skips that many number of frames specified by $a^*$). In the following, we describe the details of RL components.


\textbf{RL-Agent Components} Figure~\ref{fig:agent-training} depicts the operation of each of the agent components at training. We discuss the design of the reinforcement learning-based agent as follows.

\textbf{State} To define the state space $\mathcal{S}$, we consider a lightweight solution to determine the state of the environment as follows. We segment each frame into 3$\times$3 chunks (a total of 9 chunks) and compute the fraction of pixel change between the current frame and the previous frame. These fractions form a vector of nine elements which we consider as the state.

\begin{algorithm}
\DontPrintSemicolon
  
    Initialize $s$\;
    \While{not done}{
            $a \xleftarrow{}$ Choose-Action($s$)\;
            $d \xleftarrow{} \mathcal{D}(f_i , f_a)$\;
            $R \xleftarrow{}$Compute-Reward($d,\theta,a$)\;
            $s' \xleftarrow{}$ Observe-Environment$(E)$\;
            $a' \xleftarrow{}$ Choose-Action($s'$)\;
            $Q(s,a) \xleftarrow{} Q(s,a) + \alpha [R + \gamma. Q(s',a')-Q(s,a)]$\;
            $s\xleftarrow{}s'$\;

  }

\caption{FrameHopper training using SARSA}
\label{algo:training}
\end{algorithm}
 
\textbf{Action} In Framehopper, an action $a$ refers to the number of frames to be skipped i.e. skip-length $k$. We choose the maximum possible skip-length to be the frame processing rate (fps) of the video stream (that means, the RL can skip 1 second worth of video at best). However, this can be flexible according to the target accuracy to be met. 
At the training stage, using SARSA, the agent will try a different number of actions for a corresponding state and update the $Q$ table accordingly. Note that we do a full exploration-based search to train FrameHopper. 

\textbf{Reward} As FrameHopper tries to minimize the processing rate as well as the error rate as shown in Eqn~\ref{eqn:1}, the higher the value of $k$ with lower error should get the higher reward. Considering this, we design a simple reward function $R$ that will be higher for higher $k$ only when the error due to the skipping remains bounded with a pre-specified threshold $\theta$. If the error exceeds that bar, the reward is revoked and incurred as a penalty (negative reward). This negative reward is enforced so that the agent favors skipping fewer frames in situations where there are higher chances of introducing higher errors. In that, the agent may process more frames occasionally but would always try to keep within the limit of the error threshold. The reward function is defined as follows.

\[
    R_{t}(k) = \begin{cases}
    \psi_1 (k + 1)  &, error \le \theta\\
    - \psi_2 k  &, error>\theta\\
    \end{cases}
\]
\begin{algorithm}
  \DontPrintSemicolon
  \SetKwFunction{FMain}{Compute-Reward}
  \SetKwProg{Pn}{Function}{:}{}
  \Pn{\FMain{$d$, $\theta$, $k$}}{
        \If{$d \le \theta$}{
                $R \xleftarrow{} \psi_1 (k + 1) $\;
            }
   		\Else{
   		       $R \xleftarrow{} - \psi_2 k$\;
   		}
        \KwRet $R$\;
    }
\SetKwFunction{FMain}{Observe-Environment}
  \SetKwProg{Pn}{Function}{:}{}
  \Pn{\FMain{$f_{prev}$, $f_{next}$}}{
        $\mathbf{d} \xleftarrow{} \bigcup_{i=1}^{9}\mathcal{D}_{pixel}(f_i^{prev},f_i^{next})$\;
        $\mathcal{M} \xleftarrow{}$ trained cluster model\;
        $s \xleftarrow{}$ getState$(\mathcal{M}, \mathbf{d})$\;
        \KwRet $s$\;
        
    }
\SetKwFunction{FMain}{Choose-Action}
  \SetKwProg{Pn}{Function}{:}{}
  \Pn{\FMain{$s$}}{
        \KwRet $\arg\max_{a}(Q(s,a))$\;
    }
\SetKwFunction{FMain}{Perform-Action}
  \SetKwProg{Pn}{Function}{:}{}
  \Pn{\FMain{$k$}}{
        $f_{next} \xleftarrow{} f_{prev + k}$\;
        $prev \xleftarrow{} prev + k$\;
        \KwRet \;
    }
 \caption{Function of FrameHopper Agent}
 \label{algo:agent-function}
\end{algorithm}

\textbf{FrameHopper Training} In algorithm~\ref{algo:training}, we present the off-policy FrameHopper training algorithm. 
For each epoch, we compute the frame difference vector to get a state ($s_t$) and deliberately take an action $a_t$ from the action set, observe the environment ($s_{t+1}$), and compute the reward $R$. Based on these values, we update the $Q$ table entry for the current state-action pair as per the classical SARSA update rule~\cite{sutton1998introduction}.
\[
    Q(s_t,a_t) \xleftarrow{} Q(s_t,a_t) + \alpha [R_{t+1} + \gamma. Q(s_{t+1},a_{t+1})-Q(s_t,a_t)]
    \label{eqn:update}
\]

Note that we only update the $Q$ table during training and we do not take any action based on the $Q$ table. The overall training pipeline of FrameHopper is shown in Figure~\ref{fig:agent-training}. During training of FrameHopper, the states are clustered as shown in Figure~\ref{fig:agent-training}. We choose 10 clusters and fit the vectors in a partial manner using batch size 30 (minibatch K-Means) for each video segment of 3s. So, the total number of states will be 10.


\textbf{FrameHopper Variants}
For effective verification of FrameHopper, we adopt the whole frame difference (FHop-diff) rather than chunks as state vectors and follow the same procedure. The other version is getting detection results as the input of a state. In that case, the bounding box accuracy and count accuracy is used directly as input vector and followed by clustering in the same way although this method is very expensive as it needs detection results every time to get a state.  
Here, c(.) is the number of objects in a frame. $d_1 = \beta_1 \times \frac{|c_{prev} - c_{next}|} {\max(c_{prev}, c_{next})}$ is the change in the count and $d_2 = \beta_2 \times (1 - F1)$ is the change in F1 score. Both of the detection-based results are used to detect a state from model $\mathcal{M}$, $s = getState(\mathcal{M}, d_1, d_2)$.
In section~\ref{sec:results}, we evaluate only on lightweight variants of FrameHopper and observe the effectiveness of our proposed lightweight method.
\begin{figure}
    \centering
    \includegraphics[width=\linewidth]{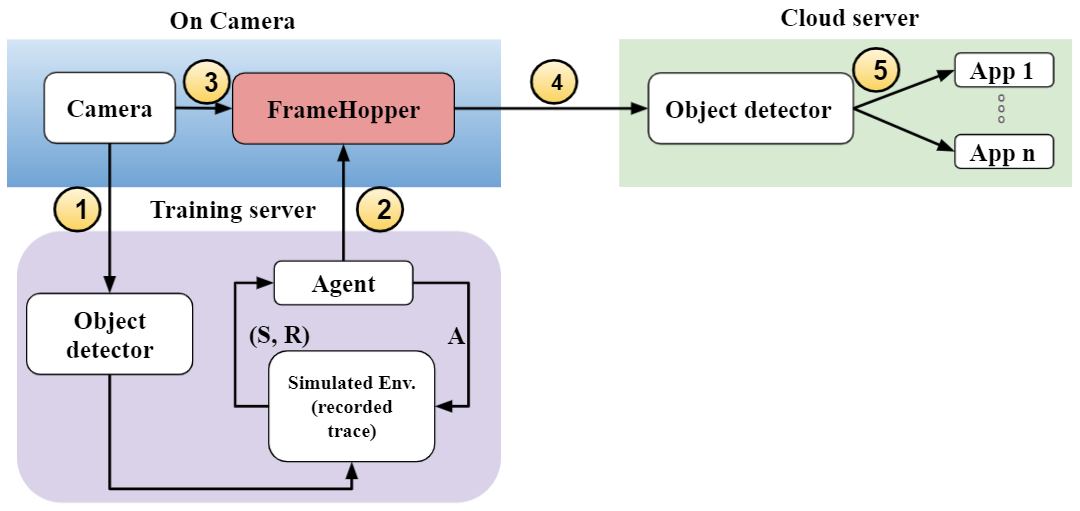}
    \caption{Framehopper task execution pipeline.}
    \label{fig:fhop-overview}
\end{figure}

\textbf{FrameHopper Inference}
As we propose an off-policy-based RL-Agent training, the agent will collect frame features to compute the state from clusters and choose skip-length $k$ as action according to the cluster at inference time. FrameHopper inference algorithm is shown in Algorithm~\ref{algo:inference}.
\begin{algorithm}
\DontPrintSemicolon

    Initialize reference frame $f_{prev}$\;
    $Agent \xleftarrow{}$ Trained RL-Agent\;
    \For{each processed frame $f_i$}{
    
            $s \xleftarrow{}$ $Agent$.Observe-Environment$(f_{i},f_{prev})$\;
            $a \xleftarrow{}$ $Agent$.Choose-Action$(s)$\;
            $Agent$.Perform-Action$(a)$\;
  }


\caption{Framehopper Inference}
\label{algo:inference}
\end{algorithm}

\subsection{System Design of FrameHopper}
In Figure~\ref{fig:fhop-overview}, we show how FrameHopper can be coupled with a camera running in an edge device. The task execution pipeline of FrameHopper is as follows.
\begin{figure}
        \subfigure[JK-1]{
        \includegraphics[width=0.3\linewidth]{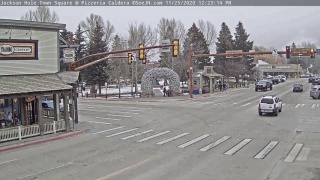}
        }%
        \subfigure[TV-2]{
               \includegraphics[width=0.3\linewidth]{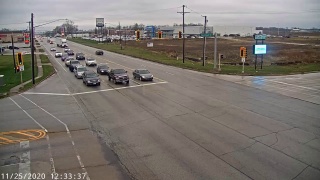}
        }%
        \subfigure[KC-1]{
            \includegraphics[width=0.3\linewidth]{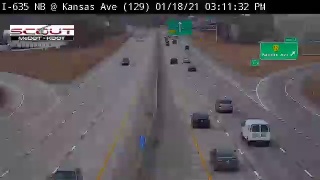}
        }
        \caption{Sample video frames}
        \label{fig:sample-videos}
\end{figure}
\Circled{1} To train the Framehopper, frames are passed to a training server that will be close to the edge. The RL-Agent will be trained using an object detector for annotation and fraction of pixel changed in a frame. After the agent is trained, \Circled{2} the agent will be deployed at the edge for frame filtering. \Circled{3} Incoming frames from the camera will be filtered by Framehopper and \Circled{4} The selected frames by FrameHopper will be offloaded to the cloud for end detection-based applications \Circled{5} running in the cloud server. The advantage of this setup is that we can filter frames directly from the camera feed which helps to reduce the communication cost.
\section{Evaluation}
\subsection{Datasets}
We run the experiments on a variety of videos with different time slots in a day and of different resolutions. There are a total of seven video datasets collected from youtube streaming i.e. Jackson~\cite{Jackson} and Tilton Village~\cite{tilton} and from KCScout~\cite{kc} camera comprising a total of around 43 min video. The duration of each video is between 5-13 min. A subset of video scenes is shown in Figure~\ref{fig:sample-videos}. All the videos have 30 FPS (frames per second). A detailed description of each dataset is shown in Table~\ref{tab:Dataset}.

\begin{table}[]
\caption{Description of Datasets}
\label{tab:Dataset}
    \centering
    \adjustbox{width=0.95\linewidth}{
    \begin{tabular}{cccccc}
    \toprule[2pt]
     \textsc{Dataset} & \textsc{Video}&\textsc{Duration} & \textsc{Resolution}& \textsc{Start} & \textsc{Total} \\
     & & & \textsc{Rate} & \textsc{Time} &\textsc{Frames} \\
    \midrule[1pt]
    \textsc{Jackson} & JK-1& 5.03 min & 1920 $\times$ 1080 & 01:23 PM & 9055\\
    &JK-2 & 5.08 min & 1920 $\times$ 1080 & 12:19 PM& 9161\\
    &JK-3 & 5.32 min & 1920 $\times$ 1080 & 04:48 PM & 9584\\
     \midrule[1pt]
    \textsc{Tilton} & TV-1& 5 min & 1280 $\times$ 720&  01:29 PM & 8987\\
    \textsc{Village}&TV-2& 5.33 min & 1280 $\times$ 720&  12:34 PM & 9587\\
    &TV-3& 5.5 min & 1280 $\times$ 720&  03:27 PM & 9888\\
    \midrule[1pt]
    \textsc{KCScout} & KC-1 &12.13 min  &640 $\times$ 360 & 03:09 PM & 21864\\
    
    \bottomrule[2pt]
    \end{tabular}
}
\end{table}
\subsection{Performance Metric}
In this section at first we discuss a formal definition of each of the metric that are used to evaluate FrameHopper. The metrics with their formulation is shown in Table~\ref{tab:per-metric}.

\begin{table}[!htbp]
\caption{Performance Metric}
    \label{tab:per-metric}
    \centering
    \resizebox{0.6\linewidth}{!}{%
    \begin{tabular}{l|c}
    \toprule[1.2pt]
    Total number of frames & $|N|$\\
    \midrule
        Number of frames processed & $|P|$\\
        \midrule
        Fraction of processed frames & $P_\theta = \frac{|P|}{|N|}$ \\
        \midrule
        Fraction of frames filtered & $1 - \frac{|P|}{|N|}$\\
        \midrule
        Error due to skip & $E_\theta = \frac{\sum_{i\in P} \mathcal{D}(i,\kappa(i))}{|N| - |P|}$\\
        \midrule
        Bounding box accuracy & $F1$ score\\
    \bottomrule[1.2pt]
    \end{tabular}}
\end{table}
The error due to skipping of frames depends on a certain threshold $(\theta = 1 - Accuracy_{target})$. In most cases, we use bounding box accuracy F1 as target accuracy $(\theta = 1 - F1)$. We use $\theta$ and target F1 interchangeably throughout the paper. As the frames are filtered out at the edge, the earlier processed frame is a representative of the next skipped frames to the cloud, so the error $E_\theta$ will be the accumulation of distance between the reference processed frames and the next skipped frames. The distance measure is discussed in Section~\ref{sec:framehopper}. The formal definition is included in Table~\ref{tab:per-metric}. As the error is occurred due to skipping, we can call it an error per skipped frame from error formulation as indicated in Table~\ref{tab:per-metric}.\par 
\textbf{Oracle solution} For achieving a target accuracy, The oracle solution exactly defines which frame to be processed or skipped. Thus, for a sequence of video frames, the oracle solution produces a fraction of processing frames and errors due to skipping which will be minimal than any solution.\par
\textbf{Pseudo ground truth generation}
Manual annotation of each video frame for object detection is very expensive and infeasible for real-time processing. As a result, we select an object detection model and consider its output as the pseudo ground truth annotation for the video frame. We use EfficientDet-D4~\cite{tan2020efficientdet}, a moderate object detection model from EfficientDet series for video frame annotation. Due to lacking of actual ground truth, unavoidable error may occur which results in a slight change of the oracle solution. It is to be noted that no significant failure is observed during experiments due to taking pseudo ground truth annotation of the frames.
\subsection{Results}\label{sec:results}
We have experimented with two variants of FrameHopper i.e. FHop-diff (pixel difference of whole frames), FHop-slice (pixel difference between constant number of respective chunks between two frames),  Reducto~\cite{reducto} and the oracle solution and compared bounding box accuracy, count accuracy, and fraction of processed frame on each video file. To evaluate the proposed approaches, we have considered 50\% of the frames of each video as a training set and the rest as a test dataset for each scheme for a fair comparison.
At the time of training, the target bounding box accuracy is 0.7, 0.8, and 0.9.
\begin{figure}
        \subfigure[]{
                \includegraphics[width=0.45\linewidth]{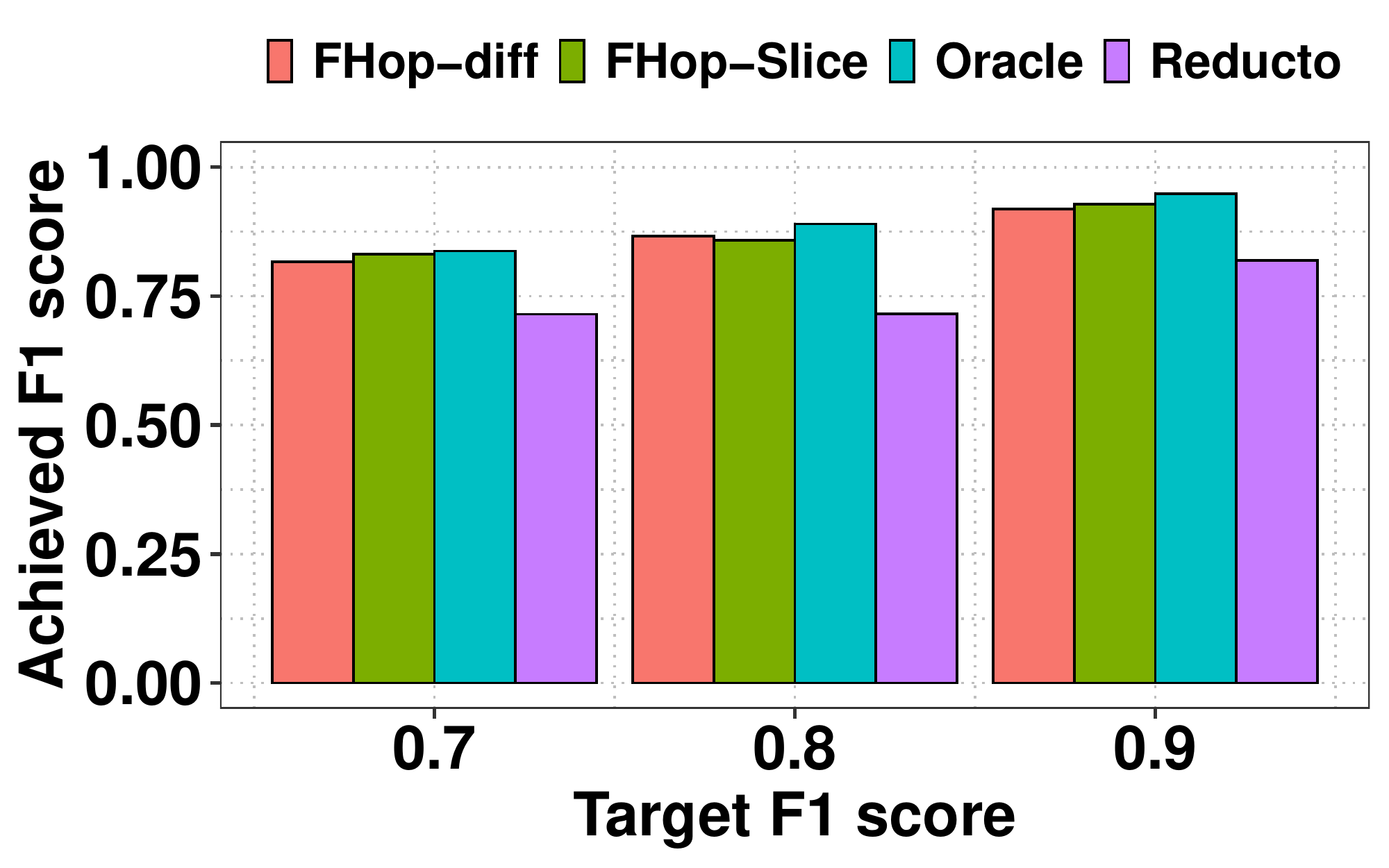}
        }%
        \subfigure[]{
                \includegraphics[width=0.45\linewidth]{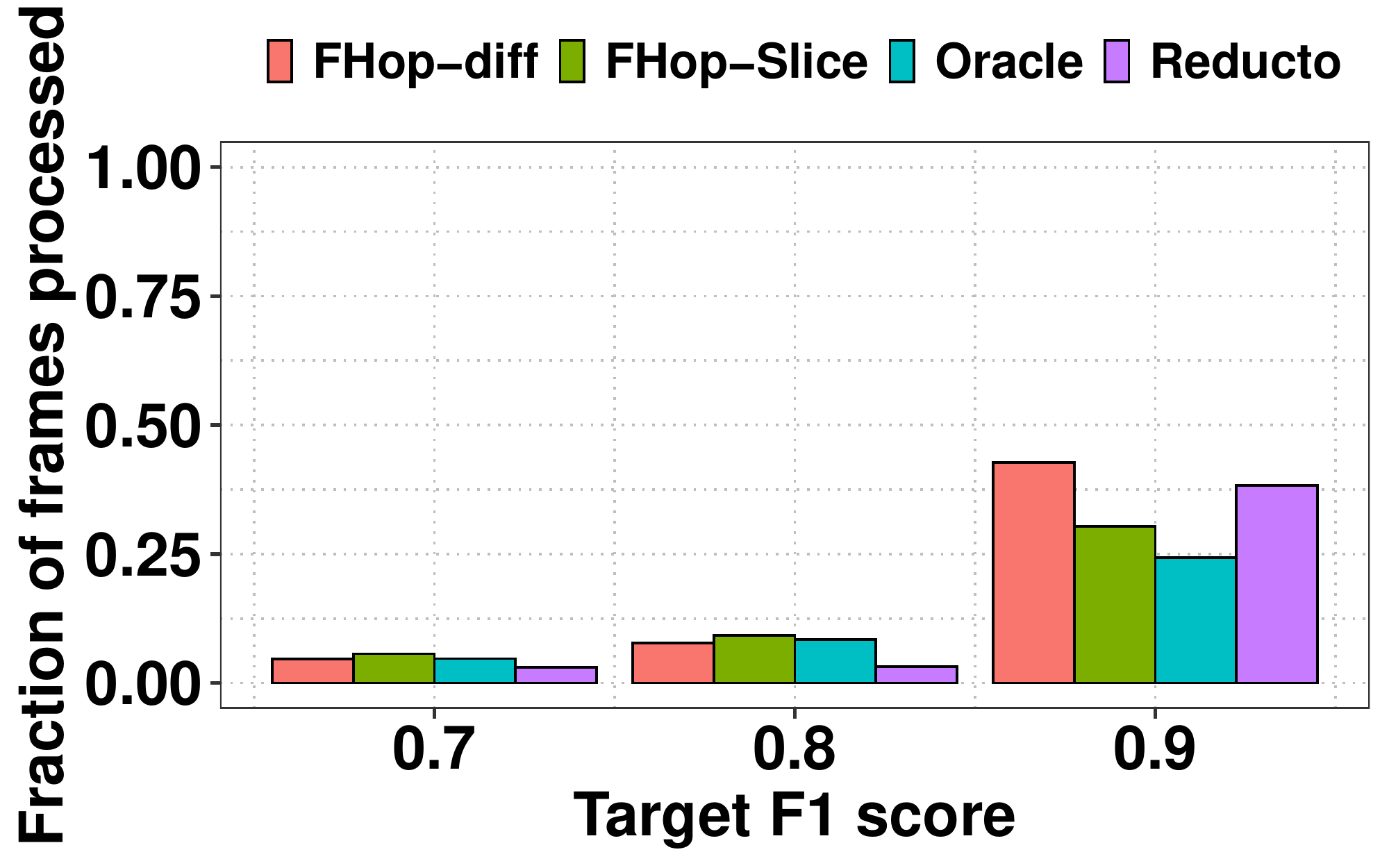}
        }
        \subfigure[]{\includegraphics[width=0.45\linewidth]{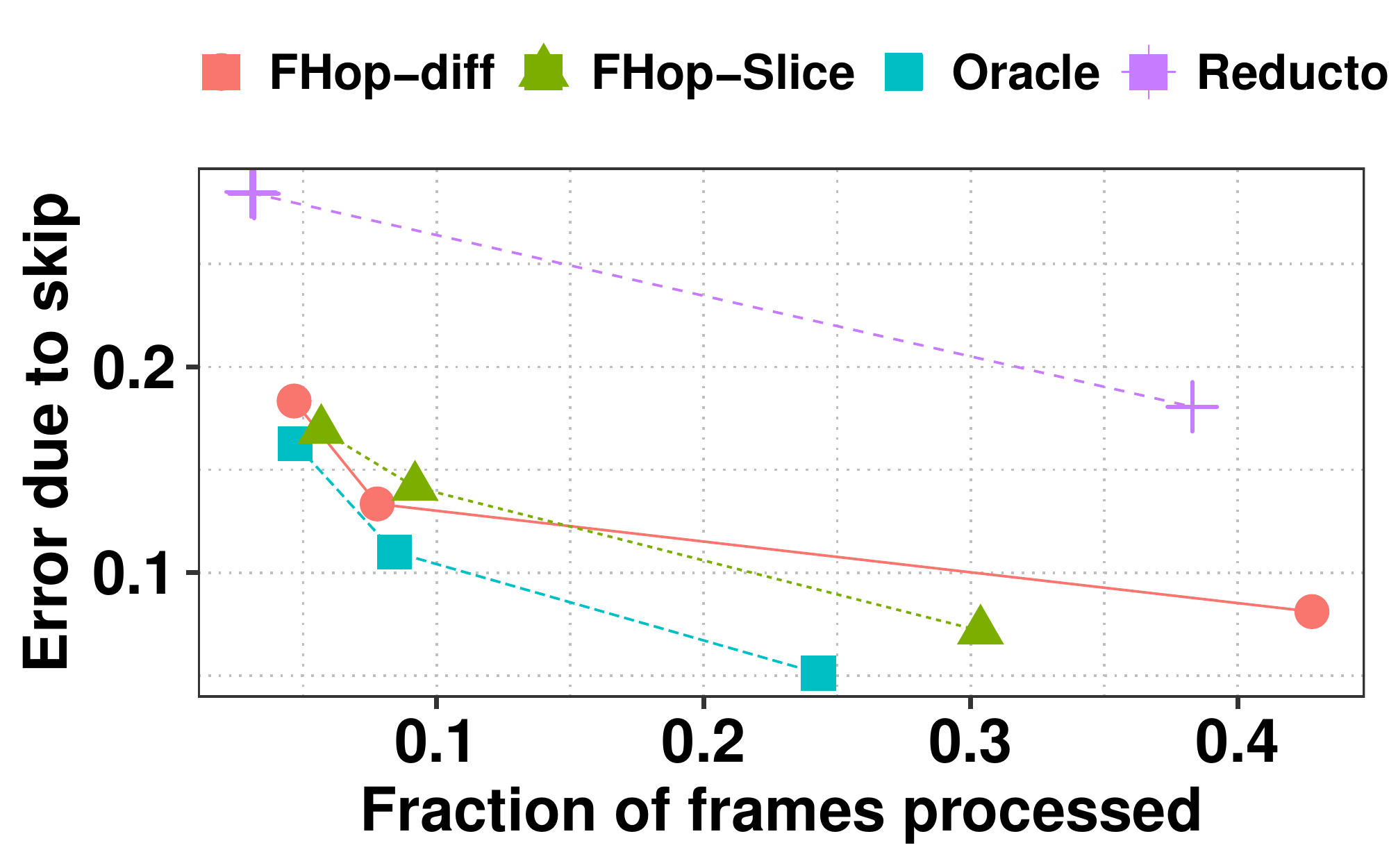}
        }%
        \subfigure[]{\includegraphics[width=0.45\linewidth]{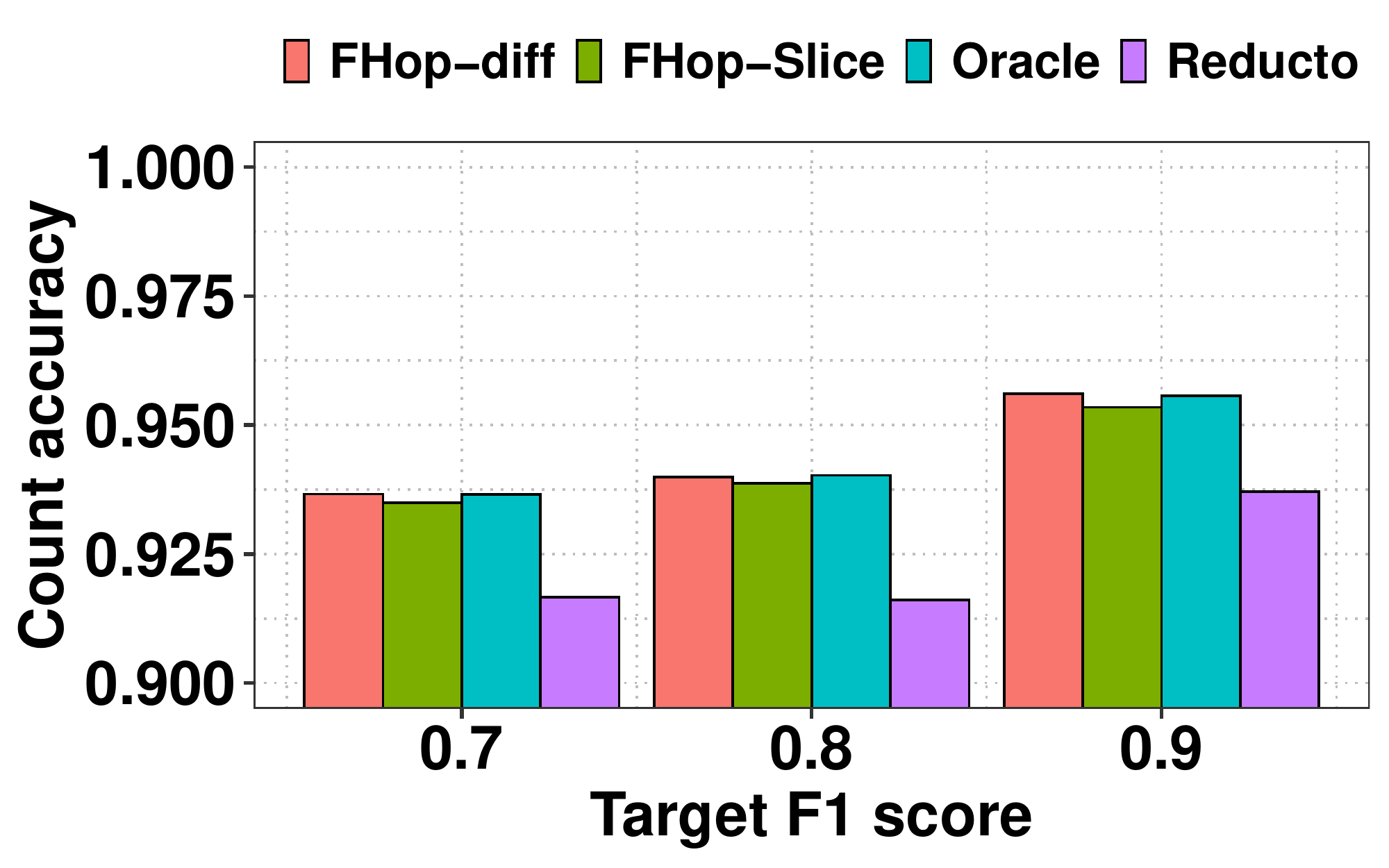}
        }
        
        \caption{ (a) Achieved F1 score vs target F1 score (b) Fraction of frame processed to achieve the target accuracy (c) Error due to skipping vs fraction of frames processed by each scheme. (d) Count accuracy for target F1 score. The FrameHopper acts closer to the oracle solution than Reducto.}
        \label{fig:jackson-all-4}
\end{figure}
\begin{figure}
        \subfigure[]{
                \includegraphics[width=0.45\linewidth]{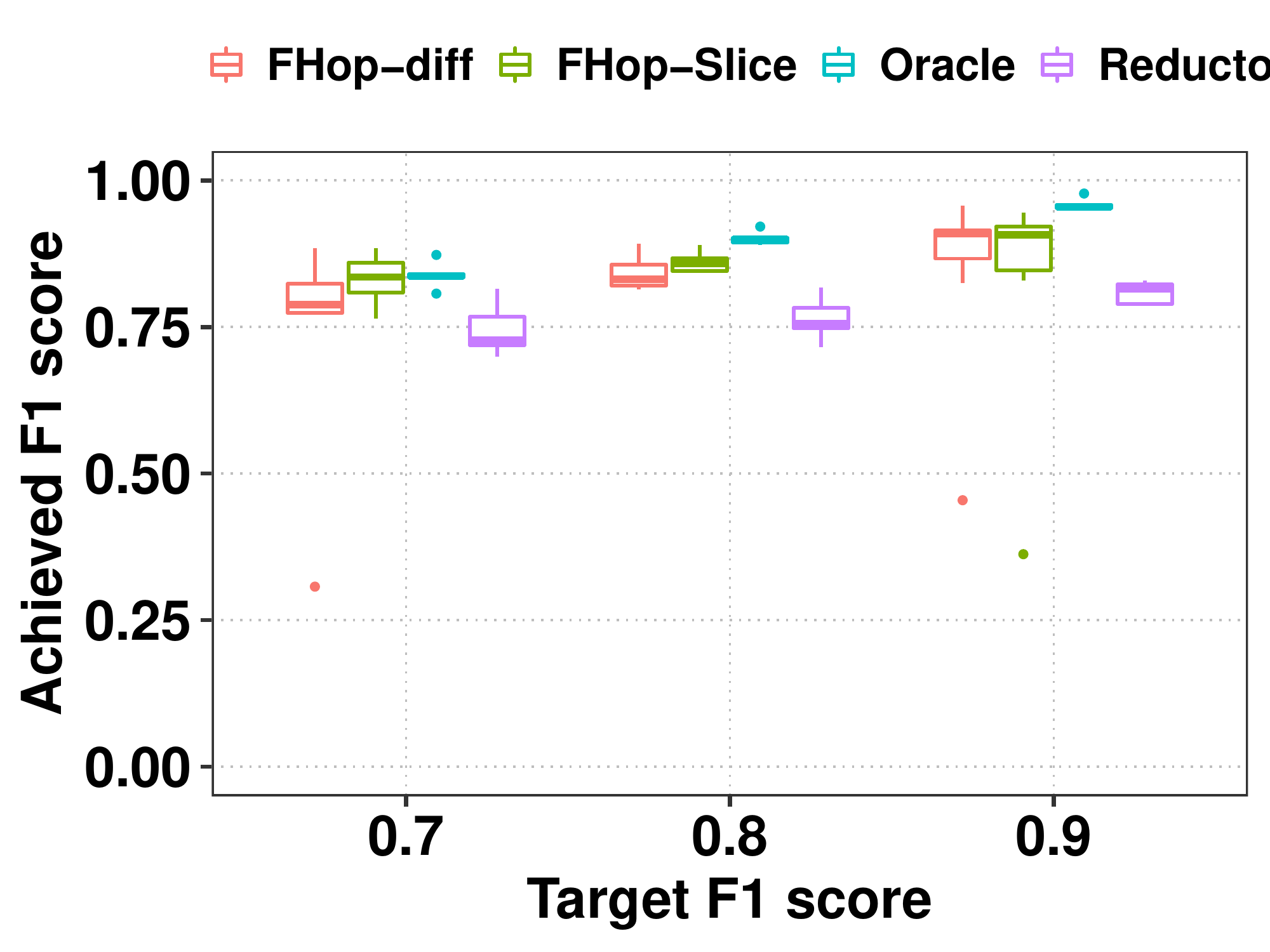}
        }%
        \subfigure[]{
                \includegraphics[width=0.45\linewidth]{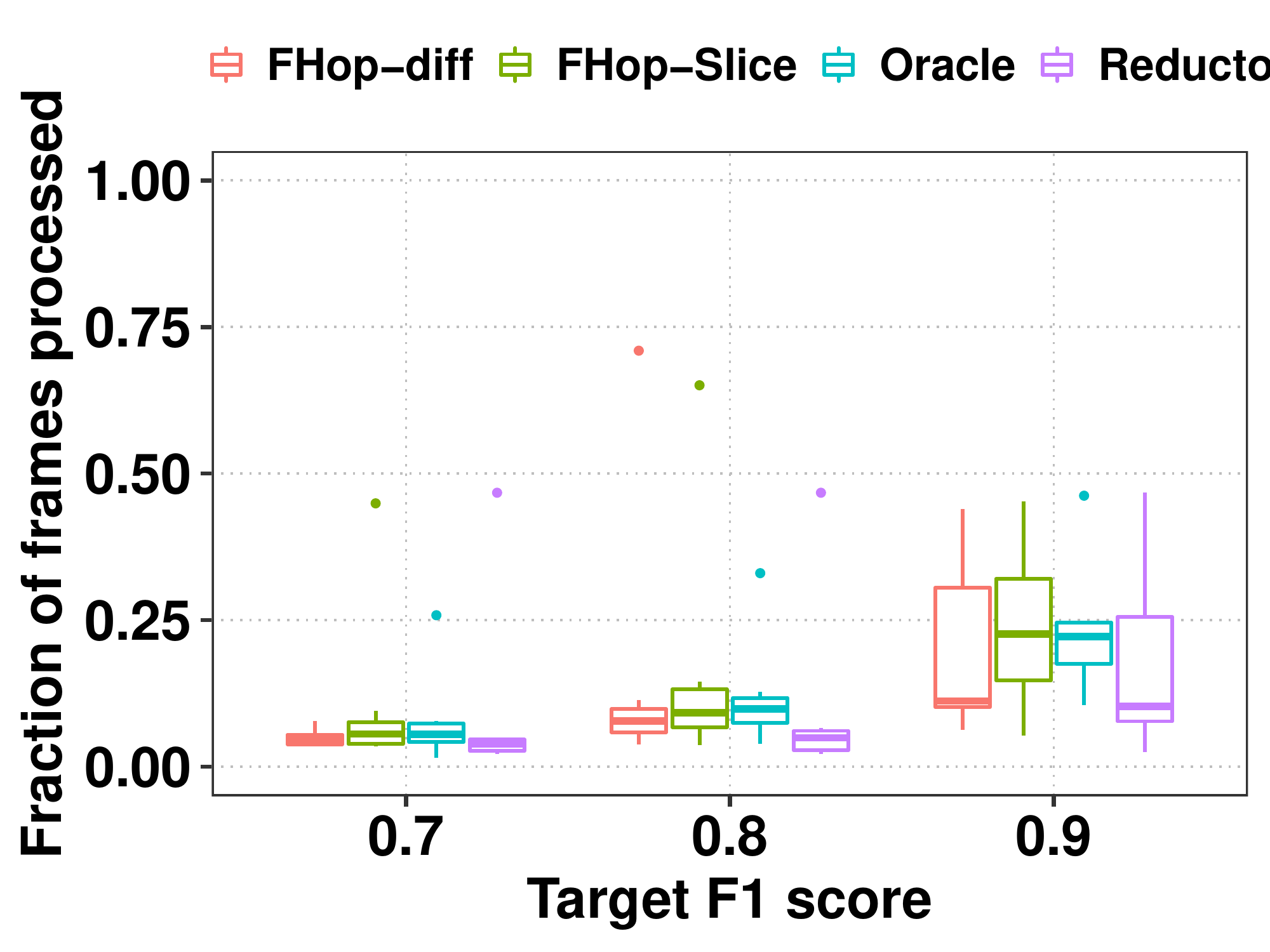}
        }
        \subfigure[]{
                \includegraphics[width=0.45\linewidth]{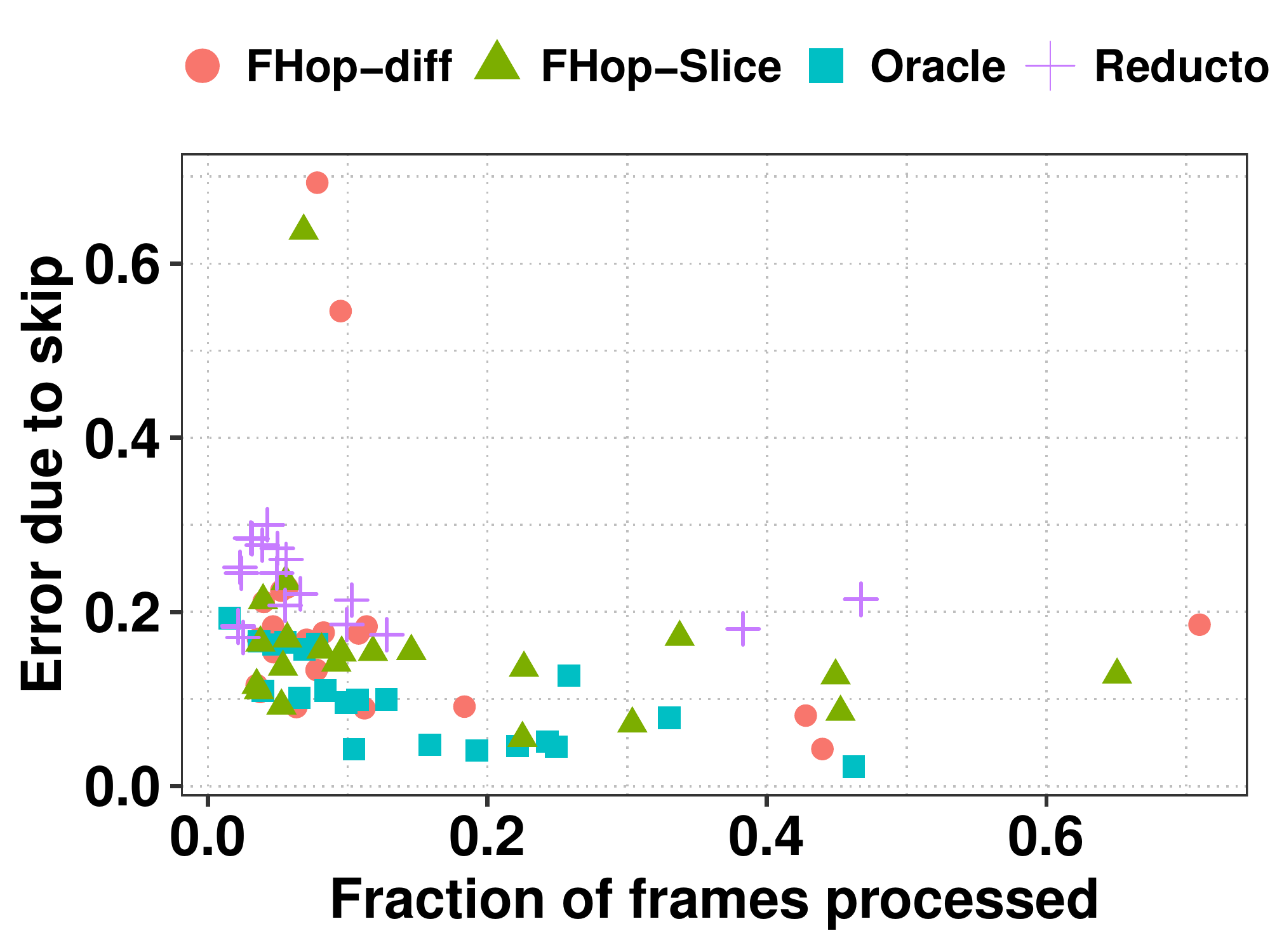}
        }%
        \subfigure[]{\includegraphics[width=0.45\linewidth]{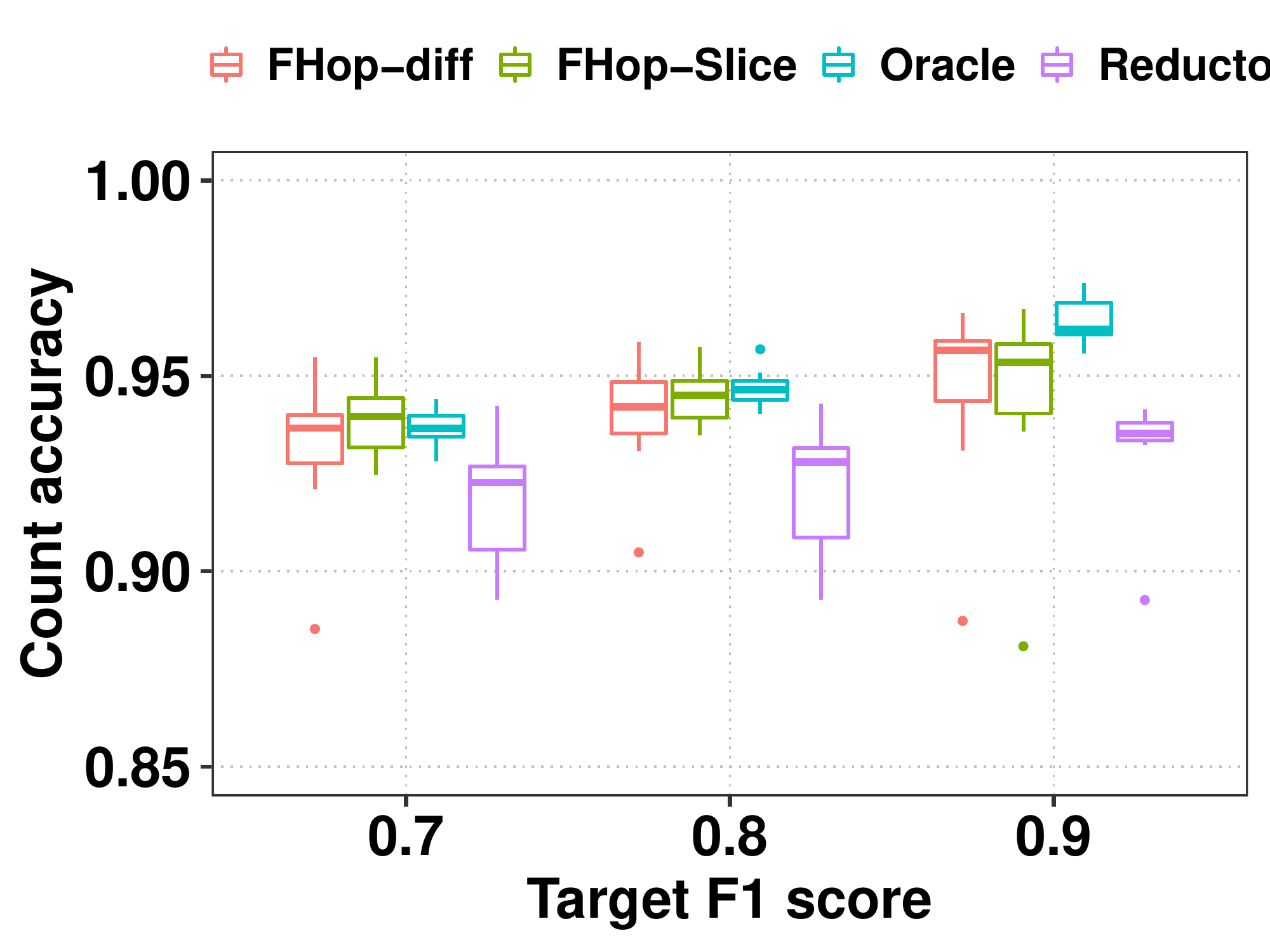}
        }
        \caption{(a) Achieved F1 score vs target F1 score (b) Fraction of frame processed vs target F1 (c) Error due to skipping vs fraction of frames processed by each scheme. (d) Count accuracy change due to target F1.}
        \label{fig:all-results}
\end{figure}
For each target F1 score i.e. target accuracy, we trained each scheme on the training set and observe the frames skipped during the testing phase. After that. for a target bounding box accuracy (F1), we computed the achieved F1 score, fractions of the processed frame, and how much error is occurred due to skipping of frames for each frame, and count accuracy (as described in~\cite{reducto} as F1 score implicitly indicates the count accuracy change). Such results for one video dataset JS-1 are shown in Figure~\ref{fig:jackson-all-4}. 
\begin{figure}
    \centering
    \includegraphics[width=0.9\linewidth]{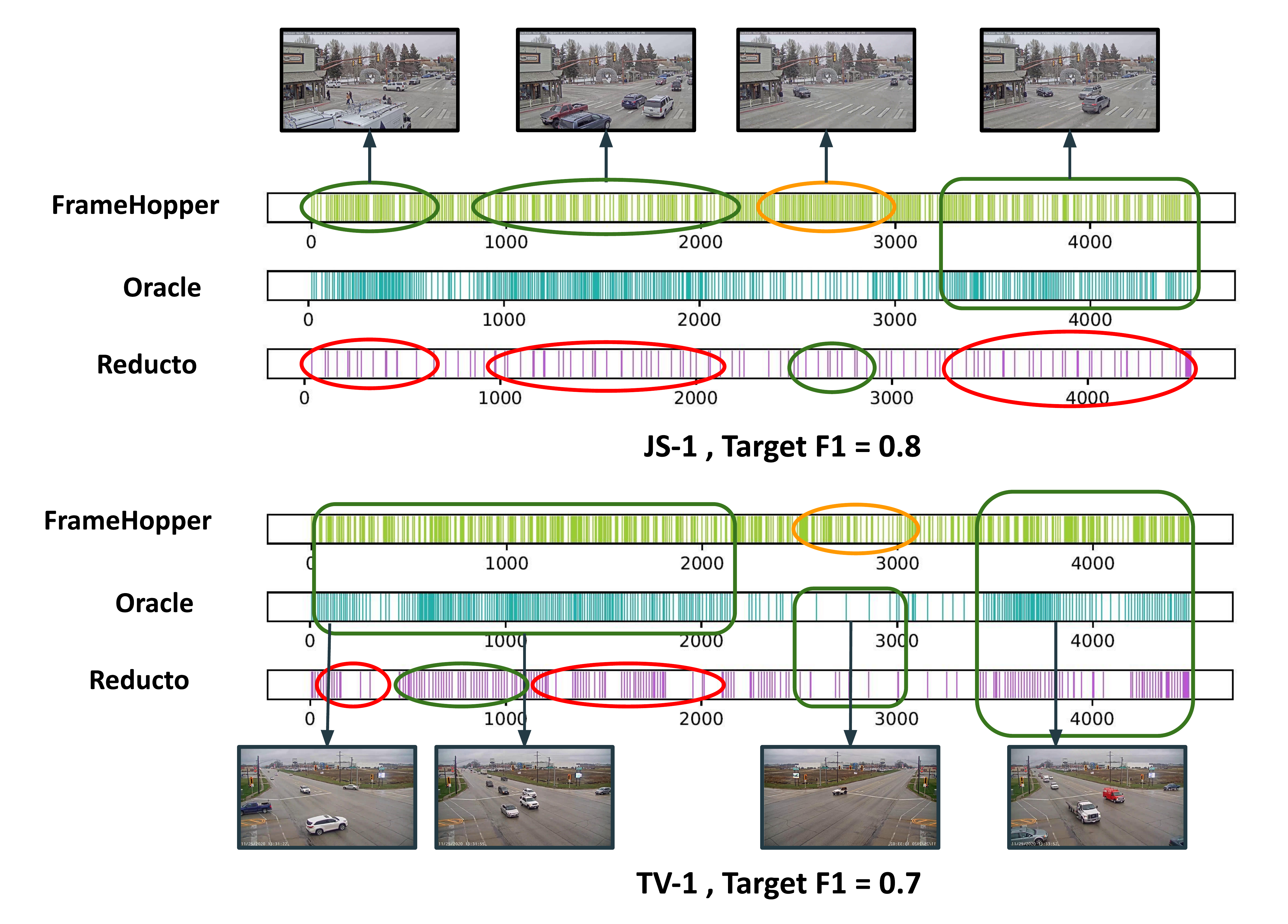}
    \caption{Qualitative comparison between FramHopper and Reducto against the oracle solution on processing frames for 2 video test datasets on different target bounding box accuracy. Green ovals indicates the overlapping of processing frames between each scheme with the oracle solution, red indicates lower processing frames and yellow indicates slightly higher processing frames.}
    \label{fig:qualitative}
\end{figure}
 \begin{figure}
        \subfigure[JS-1]{\includegraphics[width=0.44\linewidth]{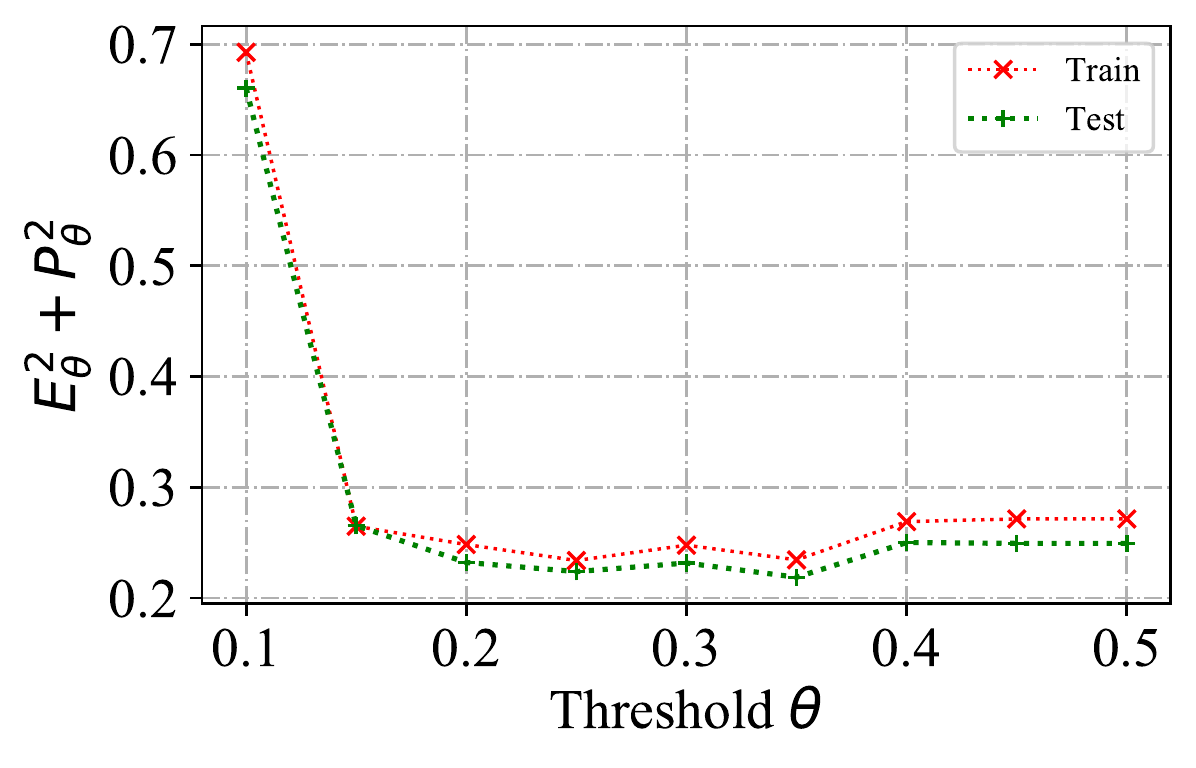}
        }%
        \subfigure[TV-1]{
                \includegraphics[width=0.45\linewidth]{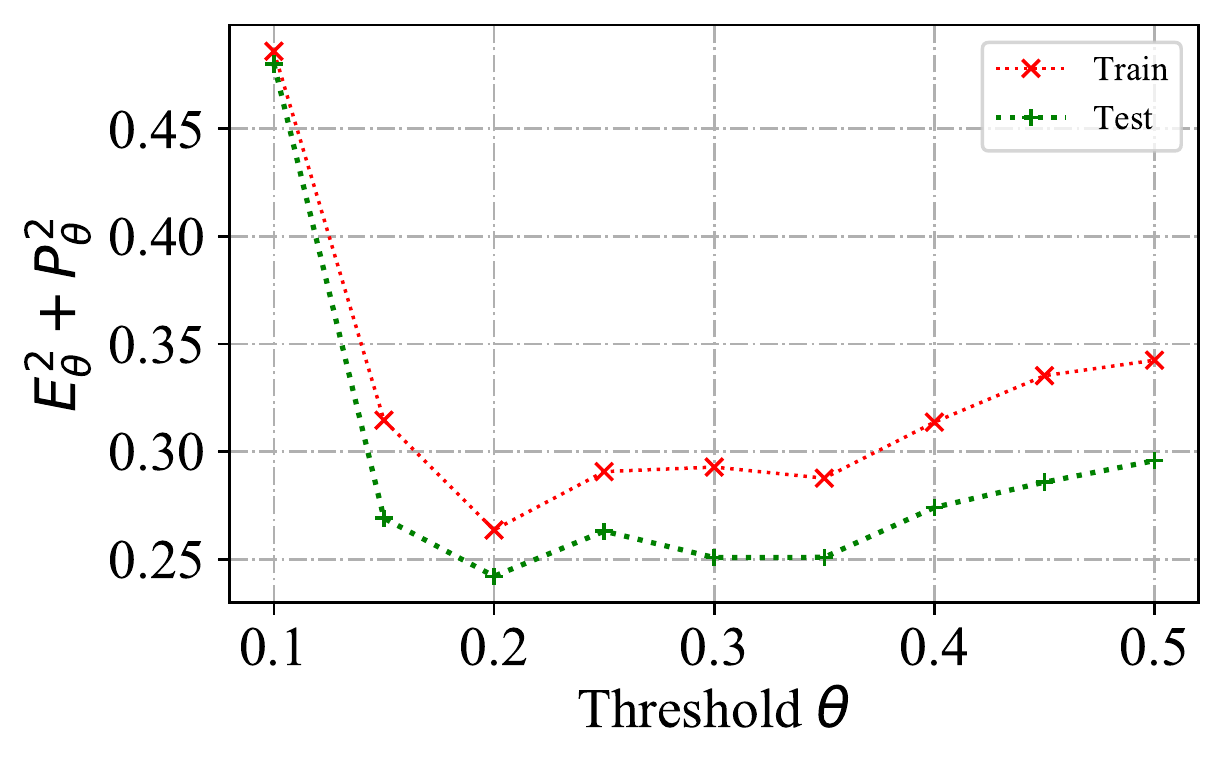}
        } 
        \caption{Selection of threshold based on minimal error and minimal processing of frames.}
    \label{fig:optimal-threshold-finding}
\end{figure}
In Figure~\ref{fig:jackson-all-4}, to meet the F1 score of 0.9, we observe the trivial scenario i.e. fraction of processed frames are higher than other low target F1 score for all schemes. It is also evident that for a higher F1 score, as the frames are skipped less, the count accuracy and achieved F1 score is higher with a higher fraction of processed frames. We compare Reducto with FrameHopper keeping the oracle solution as a baseline and measuring different performance metrics. Reducto appears to filter more and due to that sometimes it fails to achieve target accuracy. On the contrary, FrameHopper favors accuracy and always catch up the target accuracy, albeit slightly higher number of frames being selected and processed.\par 
We identify FrameHopper to be closer to the oracle result than Reducto which dictates the supremacy of FrameHopper over Reducto. To validate our point, we measure the same performance matrices for each scheme in all seven video datasets. We have observed the similar characteristics of FrameHopper of being closer to the oracle solution. The results are shown in boxplot in Figure~\ref{fig:all-results}. It is to be noted that the main FramHopper is denoted by FHop-slice in the figures and the FHop-diff a variant of FrameHopper, is denoted where pixel difference between two frames is taken into account only rather than a pixel difference of frame chunks. It can be easily identified from Figure~\ref{fig:all-results} that FHop-slice performs better than FHop-diff variant. The plot of error due to skipping vs the fraction of processed frames of various filtering methods (Figure~\ref{fig:all-results} d) indicates the high efficiency of the methods if the error is lower for lower frame processing i.e. higher frame skipping. Here, FrameHopper follows the oracle solution whereas the error of Reducto is higher. A qualitative comparison is shown in Figure~\ref{fig:qualitative} among FrameHopper, Oracle, and Reducto schemes on processed frames of test datasets where FrameHopper performs better than Reducto for higher target accuracy. For lower target accuracy, FrameHopper matches with oracle solution over a wide range. For a small range, Reducto is slightly better than FrameHopper in eliminating redundant frames for achieving lower target accuracy.\\
\textbf{Application agnostic threshold selection}
We discuss a scenario where the threshold or target accuracy will be chosen at training based on the current video scene. We consider the bounding box accuracy i.e. $F1$ score to be the one for the application-agnostic scene as it takes care of object location and labels both. Generally, the target threshold will be chosen where the $|P|$ is lower and the $E(\theta)$ is lower. At the profiling step (training), for a range of $\theta$ between 0.1-0.5 with an interval of 0.05, the best $\theta$ will be the one which minimizes ($P(\theta)$) as well as the, $E_\theta$. The optimization problem can be formulated as follows.
\begin{mini!}|l|[3]
{E_{\theta}, P_{\theta}}{E_{\theta}^2 + P_{\theta}^2}{}{}
\label{eqn:optimal-theta}
\end{mini!}

Hence, for a specific video, the optimal $\theta$ will be where the $E_{\theta}^2 + P_{\theta}^2$ value is minimum. This method might be used in a scenario where a threshold is not specified by the user. Instead, a suitable threshold $\theta$ is asked for a minimal amount of processing with minimal error.\par  
\begin{figure}
        \subfigure[]{\includegraphics[width=0.45\linewidth]{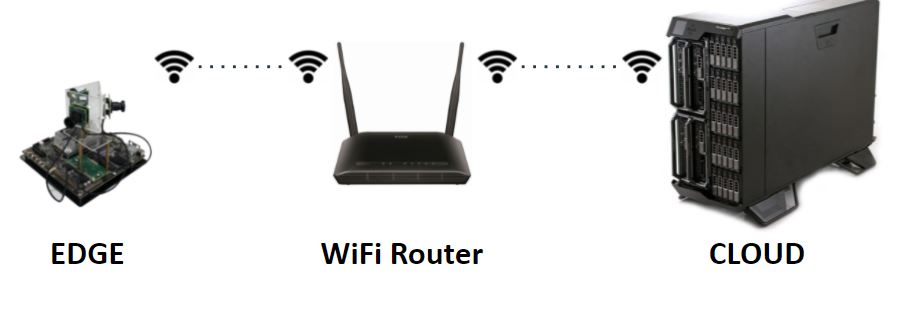}
        }%
        \subfigure[]{
                \includegraphics[width=0.4\linewidth]{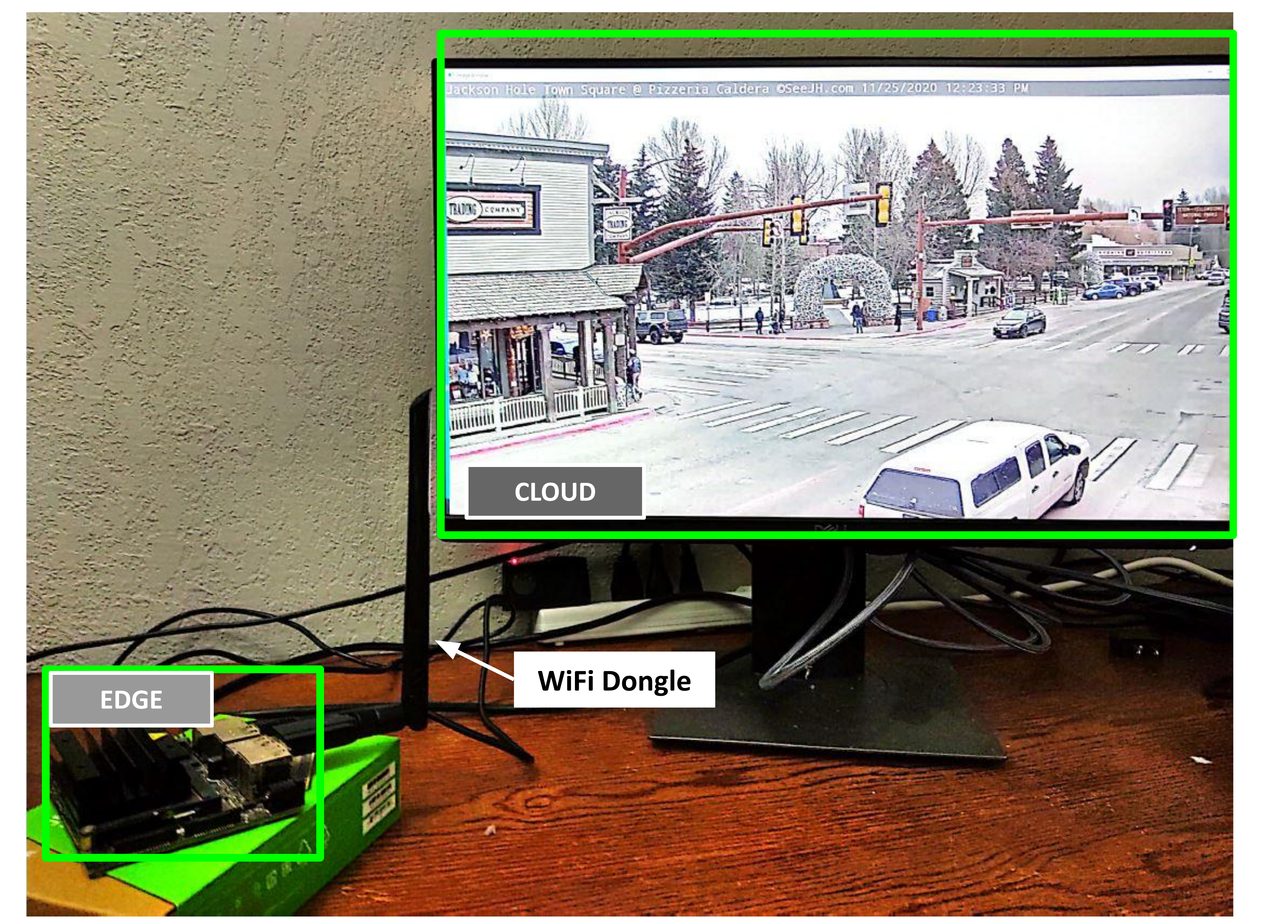}
        }
        \caption{System Deployment of FrameHopper}
        \label{fig:deploy}
\end{figure}

\begin{table}[!htbp]
\caption{Video Processing Speed}
\label{tab:fps}
\centering
\resizebox{0.9\linewidth}{!}{%
    \begin{tabular}{ccccc}
    \toprule[1.2pt]
     Method & Accuracy & Fraction   & Edge Processing & Speed\\
     &&Filtered (\%) & Speed (fps) &Improvement\\
     
    \midrule[1pt]
    Baseline & 100.00 & 0.00 & 16.75 & 1$\times$\\
    \midrule
    Reducto & 71.58 & 96.85 &50.11 & 3$\times$\\
    \midrule
    Framhopper & 85.83 & 90.82 & \textbf{63.49} & \textbf{4$\times$}\\
    \bottomrule[1.2pt]
    \end{tabular}
}
\end{table}

\textbf{Prototype Implementation}
We implement our approach in an edge-cloud setup. As the edge node of our setup, we use Jetson Nano and as cloud node, we use a workstation having NVIDIA Quadro
RTX 4000 GPU and i7-9750 CPU@2.60GHz as shown in Figure~\ref{fig:deploy}. We deploy the RL agent at the edge.

After processing the last frame, the agent determines its state and calculates the skip-length, $k$ as its action and skips the next $k$ frames. We select a video from Jackson [JS-1] and run it on Jetson Nano for Reducto, FrameHopper, and baseline scheme. The baseline scheme is sending each frame to the cloud. While sending the frames, we observe a decrease in FPS i.e. from 30 to  16.75 since a preprocessing cost such as frame read, frame encoding is included to send the frame. For the same setup, we ran Reducto and FramHopper and observe a gain in processing speed at edge for the target bounding box accuracy 0.8. Table~\ref{tab:fps} indicates the supremacy of FrameHopper over Reducto. FrameHopper do not need to read each frame or process it to identify which frames to filter, rather it just waits according to the skip-length to process the next future frame. As a result, the edge processing speed increases compared to Reducto.
\section{Conclusion}
In this paper, we present FrameHopper, a lightweight reinforcement learning-based solution for edge devices. We have also discussed other variations of FrameHopper. In the context of accuracy and filtering, FrameHopper is closer to the oracle solution than others from recent literature. However, a constant amount of frame chunks are reported to compute the state of RL-Agent. Making the number of frame chunks dynamic will lead to a latency accuracy trade-off. So, handling multi-stream videos using dynamic chunks and one FramHopper agent will be another interesting phenomenon that will be a direction to our future research.

\bibliographystyle{IEEEtran}
\bibliography{references}

\begin{thebibliography}{10}
\providecommand{\url}[1]{#1}
\csname url@samestyle\endcsname
\providecommand{\newblock}{\relax}
\providecommand{\bibinfo}[2]{#2}
\providecommand{\BIBentrySTDinterwordspacing}{\spaceskip=0pt\relax}
\providecommand{\BIBentryALTinterwordstretchfactor}{4}
\providecommand{\BIBentryALTinterwordspacing}{\spaceskip=\fontdimen2\font plus
\BIBentryALTinterwordstretchfactor\fontdimen3\font minus
  \fontdimen4\font\relax}
\providecommand{\BIBforeignlanguage}[2]{{%
\expandafter\ifx\csname l@#1\endcsname\relax
\typeout{** WARNING: IEEEtran.bst: No hyphenation pattern has been}%
\typeout{** loaded for the language `#1'. Using the pattern for}%
\typeout{** the default language instead.}%
\else
\language=\csname l@#1\endcsname
\fi
#2}}
\providecommand{\BIBdecl}{\relax}
\BIBdecl

\bibitem{bastani2020miris}
F.~Bastani, S.~He, A.~Balasingam, K.~Gopalakrishnan, M.~Alizadeh,
  H.~Balakrishnan, M.~Cafarella, T.~Kraska, and S.~Madden, ``Miris: Fast object
  track queries in video,'' in \emph{Proceedings of the 2020 ACM SIGMOD
  International Conference on Management of Data}, 2020, pp. 1907--1921.

\bibitem{gu2020appearance}
X.~Gu, H.~Chang, B.~Ma, H.~Zhang, and X.~Chen, ``Appearance-preserving 3d
  convolution for video-based person re-identification,'' in \emph{European
  Conference on Computer Vision}.\hskip 1em plus 0.5em minus 0.4em\relax
  Springer, 2020, pp. 228--243.

\bibitem{hou2020temporal}
R.~Hou, H.~Chang, B.~Ma, S.~Shan, and X.~Chen, ``Temporal complementary
  learning for video person re-identification,'' in \emph{European Conference
  on Computer Vision}.\hskip 1em plus 0.5em minus 0.4em\relax Springer, 2020,
  pp. 388--405.

\bibitem{yang2020spatial}
J.~Yang, W.-S. Zheng, Q.~Yang, Y.-C. Chen, and Q.~Tian, ``Spatial-temporal
  graph convolutional network for video-based person re-identification,'' in
  \emph{Proceedings of the IEEE/CVF Conference on Computer Vision and Pattern
  Recognition}, 2020, pp. 3289--3299.

\bibitem{zhang2020refineface}
S.~Zhang, C.~Chi, Z.~Lei, and S.~Z. Li, ``Refineface: Refinement neural network
  for high performance face detection,'' \emph{IEEE Transactions on Pattern
  Analysis and Machine Intelligence}, 2020.

\bibitem{tan2020efficientdet}
M.~Tan, R.~Pang, and Q.~V. Le, ``Efficientdet: Scalable and efficient object
  detection,'' in \emph{Proceedings of the IEEE/CVF Conference on Computer
  Vision and Pattern Recognition}, 2020, pp. 10\,781--10\,790.

\bibitem{liu2021ntire}
J.~Liu, N.~Inkawhich, O.~Nina, and R.~Timofte, ``Ntire 2021 multi-modal aerial
  view object classification challenge,'' in \emph{Proceedings of the IEEE/CVF
  Conference on Computer Vision and Pattern Recognition}, 2021, pp. 588--595.

\bibitem{kortylewski2020combining}
A.~Kortylewski, Q.~Liu, H.~Wang, Z.~Zhang, and A.~Yuille, ``Combining
  compositional models and deep networks for robust object classification under
  occlusion,'' in \emph{Proceedings of the IEEE/CVF Winter Conference on
  Applications of Computer Vision}, 2020, pp. 1333--1341.

\bibitem{reducto}
Y.~Li, A.~Padmanabhan, P.~Zhao, Y.~Wang, G.~H. Xu, and R.~Netravali, ``Reducto:
  On-camera filtering for resource-efficient real-time video analytics,'' in
  \emph{Proc. of the Annual conf. of the ACM Special Interest Group on Data
  Comm. on the appl., technol., archit., and protocols for compute. commun.},
  2020, pp. 359--376.

\bibitem{pakha2018reinventing}
C.~Pakha, A.~Chowdhery, and J.~Jiang, ``Reinventing video streaming for
  distributed vision analytics,'' in \emph{10th $\{$USENIX$\}$ Workshop on Hot
  Topics in Cloud Computing (HotCloud 18)}, 2018.

\bibitem{kang2017optimizing}
D.~Kang, J.~Emmons, F.~Abuzaid, P.~Bailis, and M.~Zaharia, ``Optimizing deep
  cnn-based queries over video streams at scale,'' \emph{PVLDB}, 2017.

\bibitem{wang2018bandwidth}
J.~Wang, Z.~Feng, Z.~Chen, S.~George, M.~Bala, P.~Pillai, S.-W. Yang, and
  M.~Satyanarayanan, ``Bandwidth-efficient live video analytics for drones via
  edge computing,'' in \emph{2018 IEEE/ACM Symposium on Edge Computing
  (SEC)}.\hskip 1em plus 0.5em minus 0.4em\relax IEEE, 2018, pp. 159--173.

\bibitem{chen2015glimpse}
T.~Y.-H. Chen, L.~Ravindranath, S.~Deng, P.~Bahl, and H.~Balakrishnan,
  ``Glimpse: Continuous, real-time object recognition on mobile devices,'' in
  \emph{Proceedings of the 13th ACM Conference on Embedded Networked Sensor
  Systems}, 2015, pp. 155--168.

\bibitem{cai2015learning}
Z.~Cai, M.~Saberian, and N.~Vasconcelos, ``Learning complexity-aware cascades
  for deep pedestrian detection,'' in \emph{Proceedings of the IEEE
  International Conference on Computer Vision}, 2015, pp. 3361--3369.

\bibitem{hsu2020traffic}
H.-M. Hsu, Y.~Wang, and J.-N. Hwang, ``Traffic-aware multi-camera tracking of
  vehicles based on reid and camera link model,'' in \emph{Proceedings of the
  28th ACM International Conference on Multimedia}, 2020, pp. 964--972.

\bibitem{Peng_2021_WACV}
G.~Peng, B.~Pang, and C.~Lu, ``Efficient 3d video engine using frame
  redundancy,'' in \emph{Proceedings of the IEEE/CVF Winter Conference on
  Applications of Computer Vision (WACV)}, January 2021, pp. 3792--3802.

\bibitem{Bhardwaj_2019_CVPR}
S.~Bhardwaj, M.~Srinivasan, and M.~M. Khapra, ``Efficient video classification
  using fewer frames,'' in \emph{Proceedings of the IEEE/CVF Conference on
  Computer Vision and Pattern Recognition (CVPR)}, June 2019.

\bibitem{fan2018watching}
H.~Fan, Z.~Xu, L.~Zhu, C.~Yan, J.~Ge, and Y.~Yang, ``Watching a small portion
  could be as good as watching all: Towards efficient video classification,''
  in \emph{IJCAI International Joint Conference on Artificial Intelligence},
  2018.

\bibitem{trafficgorkem}
G.~Kar, S.~Jain, M.~Gruteser, F.~Bai, and R.~Govindan, ``Real-time traffic
  estimation at vehicular edge nodes,'' in \emph{Proceedings of the Second
  ACM/IEEE Symposium on Edge Computing}, ser. SEC '17.\hskip 1em plus 0.5em
  minus 0.4em\relax New York, NY, USA: Association for Computing Machinery,
  2017.

\bibitem{hamaguchi2019rare}
R.~Hamaguchi, K.~Sakurada, and R.~Nakamura, ``Rare event detection using
  disentangled representation learning,'' in \emph{Proceedings of the IEEE/CVF
  Conference on Computer Vision and Pattern Recognition}, 2019, pp. 9327--9335.

\bibitem{pang2020self}
G.~Pang, C.~Yan, C.~Shen, A.~v.~d. Hengel, and X.~Bai, ``Self-trained deep
  ordinal regression for end-to-end video anomaly detection,'' in
  \emph{Proceedings of the IEEE/CVF Conference on Computer Vision and Pattern
  Recognition}, 2020, pp. 12\,173--12\,182.

\bibitem{jiang2018chameleon}
J.~Jiang, G.~Ananthanarayanan, P.~Bodik, S.~Sen, and I.~Stoica, ``Chameleon:
  scalable adaptation of video analytics,'' in \emph{Proceedings of the 2018
  Conference of the ACM Special Interest Group on Data Communication}, 2018,
  pp. 253--266.

\bibitem{filterforward}
C.~Canel, T.~Kim, G.~Zhou, C.~Li, H.~Lim, D.~G. Andersen, M.~Kaminsky, and
  S.~Dulloor, ``Scaling video analytics on constrained edge nodes,'' in
  \emph{Proceedings of Machine Learning and Systems 2019, MLSys 2019, Stanford,
  CA, USA, March 31 - April 2, 2019}, 2019.

\bibitem{fang2018nestdnn}
B.~Fang, X.~Zeng, and M.~Zhang, ``Nestdnn: Resource-aware multi-tenant
  on-device deep learning for continuous mobile vision,'' in \emph{Proceedings
  of the 24th Annual International Conference on Mobile Computing and
  Networking}, 2018, pp. 115--127.

\bibitem{kang2017noscope}
D.~Kang, J.~Emmons, F.~Abuzaid, P.~Bailis, and M.~Zaharia, ``Noscope:
  Optimizing neural network queries over video at scale,'' \emph{Proceedings of
  the VLDB Endowment}, vol.~10, no.~11, 2017.

\bibitem{nigade2020clownfish}
V.~Nigade, L.~Wang, and H.~Bal, ``Clownfish: Edge and cloud symbiosis for video
  stream analytics,'' in \emph{ACM/IEEE Symposium on Edge Computing (SEC)},
  2020.

\bibitem{zeng2020distream}
X.~Zeng, B.~Fang, H.~Shen, and M.~Zhang, ``Distream: scaling live video
  analytics with workload-adaptive distributed edge intelligence,'' in
  \emph{Proceedings of the 18th Conference on Embedded Networked Sensor
  Systems}, 2020, pp. 409--421.

\bibitem{sutton1998introduction}
R.~S. Sutton, A.~G. Barto \emph{et~al.}, \emph{Introduction to reinforcement
  learning}.\hskip 1em plus 0.5em minus 0.4em\relax MIT press Cambridge, 1998,
  vol. 135.

\bibitem{Jackson}
\BIBentryALTinterwordspacing
Jackson hole wyoming usa town square live cam. [Online]. Available:
  \url{https://www.youtube.com/watch?v=1EiC9bvVGnk&ab_channel=SeeJacksonHole}
\BIBentrySTDinterwordspacing

\bibitem{tilton}
\BIBentryALTinterwordspacing
Village of tilton - traffic camera, usa. [Online]. Available:
  \url{https://www.youtube.com/watch?v=5_XSYlAfJZM}
\BIBentrySTDinterwordspacing

\bibitem{kc}
\BIBentryALTinterwordspacing
Kcscout. [Online]. Available: \url{http://www.kcscout.com/}
\BIBentrySTDinterwordspacing

\end{thebibliography}

\end{document}